\documentclass[conference]{IEEEtran}
\IEEEoverridecommandlockouts
\usepackage[frozencache=true, finalizecache=false, cachedir=./minted-cache]{minted} 
\usepackage{cite}
\usepackage{amsmath,amssymb,amsfonts}
\usepackage{graphicx}
\usepackage{textcomp}
\usepackage{xcolor}
\usepackage[most]{tcolorbox}
\def\BibTeX{{\rm B\kern-.05em{\sc i\kern-.025em b}\kern-.08em
    T\kern-.1667em\lower.7ex\hbox{E}\kern-.125emX}}

\usepackage{tikz}
\usetikzlibrary{matrix, arrows}
\usepackage{amsmath,amssymb}
\usepackage{amsthm}
\usepackage{mathtools}
\usepackage{xspace}
\usepackage[noend]{algorithmic}
\usepackage[ruled,vlined]{algorithm2e}
\usepackage{url}
\usepackage{makeidx}
\usepackage{enumerate}
\usepackage{epstopdf}
\usepackage{booktabs}
\usepackage{color}
\usepackage[utf8]{inputenc}
\usepackage{thm-restate}
\usepackage{scalerel,stackengine}
\usepackage[shortlabels]{enumitem}
\usepackage{xr}
\usepackage{fancyvrb}
\usepackage{xcolor}
\usepackage{bold-extra}
\usepackage[width=474.18663pt]{caption}
\usepackage{subfigure}
\usepackage[most]{tcolorbox}
\usepackage{fvextra}
\usepackage[frozencache=true, finalizecache=false, cachedir=./minted-cache]{minted} 
\usepackage{float}
\usepackage{alltt}
\usepackage{soul}
\usepackage{fancyvrb}
\usepackage{multirow}
\usepackage[final]{hyperref}
\usepackage[bottom]{footmisc}
\usepackage{array}
\newcolumntype{P}[1]{>{\centering\arraybackslash}p{#1}}
\usepackage[normalem]{ulem}
\useunder{\uline}{\ul}{}

\usepackage{listings}
\usepackage{tikz}
\usetikzlibrary{shapes,calc,positioning}

\global

\tcbset{
  aibox/.style={
    width=260.18663pt,
    top=10pt,
    colback=white,
    colframe=black,
    colbacktitle=black,
    enhanced,
    center,
    breakable,
    attach boxed title to top left={yshift=-0.1in,xshift=0.15in},
    boxed title style={boxrule=0pt,colframe=white,},
  }
}
\newtcolorbox{AIbox}[2][]{aibox,title=#2,#1}

\definecolor{aigold}{RGB}{244,210, 1} 
\definecolor{aigreen}{RGB}{210,244,211} 

\sethlcolor{aigreen}

\definecolor{aired}{RGB}{255,180,181}

\newtcbox{\mybox}[1][green]{on line,
arc=0pt,outer arc=0pt,colback=#1!10!white,colframe=#1!50!black,
boxsep=0pt,left=0pt,right=0pt,top=0pt,bottom=0pt,
boxrule=0pt,bottomrule=0pt,toprule=0pt}

\begin{document}

\title{PokerGPT: An End-to-End Lightweight \\ Solver for Multi-Player Texas Hold'em via \\ Large Language Model}

\author{    \IEEEauthorblockN{Chenghao Huang\IEEEauthorrefmark{3}, Yanbo Cao\IEEEauthorrefmark{3}, Yinlong Wen\IEEEauthorrefmark{2}, Tao Zhou\IEEEauthorrefmark{3}, and Yanru Zhang\IEEEauthorrefmark{3}\IEEEauthorrefmark{4}}
    \IEEEauthorblockA{\IEEEauthorrefmark{3}University of Electronic Science and Technology of China}
    \IEEEauthorblockA{\IEEEauthorrefmark{4}Shenzhen Institute of Advanced Study, UESTC, China
    \\ zydhjh4593@gmail.com, yanbocao@std.uestc.edu.cn, zhutou@ustc.edu, yanruzhang@uestc.edu.cn, }
    \IEEEauthorblockA{\IEEEauthorrefmark{2}Sichuan Optical Character Technology Co., Ltd
    \\ lywen@chineseocr.com}
}

\maketitle

\begin{abstract}
Poker, also known as Texas Hold'em, has always been a typical research target within imperfect information games (IIGs). IIGs have long served as a measure of artificial intelligence (AI) development. Representative prior works, such as DeepStack and Libratus heavily rely on counterfactual regret minimization (CFR) to tackle heads-up no-limit Poker. However, it is challenging for subsequent researchers to learn CFR from previous models and apply it to other real-world applications due to the expensive computational cost of CFR iterations. Additionally, CFR is difficult to apply to multi-player games due to the exponential growth of the game tree size. In this work, we introduce PokerGPT, an end-to-end solver for playing Texas Hold'em with arbitrary number of players and gaining high win rates, established on a lightweight large language model (LLM). PokerGPT only requires simple textual information of Poker games for generating decision-making advice, thus guaranteeing the convenient interaction between AI and humans. We mainly transform a set of textual records acquired from real games into prompts, and use them to fine-tune a lightweight pre-trained LLM using reinforcement learning human feedback technique. To improve fine-tuning performance, we conduct prompt engineering on raw data, including filtering useful information, selecting behaviors of players with high win rates, and further processing them into textual instruction using multiple prompt engineering techniques. Through the experiments, we demonstrate that PokerGPT outperforms previous approaches in terms of win rate, model size, training time, and response speed, indicating the great potential of LLMs in solving IIGs.
\end{abstract}


\section{Introduction}
Since the birth of the artificial intelligence (AI), intelligent game research has been fertile ground for the development and innovation of AI, and has always been an important evaluation criterion for measuring its development level. In 2016, AlphaGo defeated the human world champion in the game of Go, which was considered as an important milestone for AI solving perfect information games. However, a distinctive feature of real-world games is the uncertainty of the opponent due to incomplete information. Poker is a typical imperfect information game (IIG) that has a long history as a challenging problem for developing AI that can address hidden information~\cite{waterman1970generalization}. Poker is a superb experimental and testing platform, providing an excellent arena for the exploration and validation of foundational theories and methodologies in intelligent games.

In recent years, significant advancements have been made in optimized solutions for the game of Poker. Notably, DeepStack from the University of Alberta, Canada~\cite{moravvcik2017deepstack}, and Libratus from Carnegie Mellon University (CMU), USA, have surpassed human professionals in heads-up no-limit Poker, a 2-player form of Poker~\cite{brown2018superhuman}. The works of~\cite{brown2018depth, brown2019solving} have been further proposed to improve the computational efficiency. Subsequent to these achievements, the Carnegie Mellon-designed Pluribus demonstrated superiority by defeating a human professional in 6-player no-limit Poker~\cite{brown2019superhuman}.
These prominent AI models for Poker commonly adopt iterative Counterfactual Regret Minimization (CFR) algorithms to approximate Nash equilibrium strategies, involving initially compressing the state and action space of the hand to reduce the game tree's size, followed by iterations of the CFR algorithm over the condensed game tree. However, these methods exhibit the following drawbacks:
\begin{itemize}
    \item \textbf{Computational and storage resource consumption:} Despite model reduction, significant computational and storage resources are still required, escalating the overall computational cost. For example, in 2015, the work of~\cite{bowling2015heads} consumed 4800 CPU cores, 10.9 TB memory, and 2 months to solve heads-up limit Poker through CFR.
    \item \textbf{Difficulty in multi-player extension:} Directly extending the CFR framework to multi-player Poker environments proves to be challenging. An increase in the number of players leads to an exponential growth in the game tree size, diminishing generalization performance. For example, CMU used CFR to solve 6-player no-limit Poker and consumed 12800 CPU cores, 12.5 TB memory, and 8 days~\cite{brown2019superhuman}. The lack of flexibility make it unrealistic to use existing works in real-world games.
    \item \textbf{Loss of information:} Game tree compression unavoidably results in the loss of crucial information for decision making, such as the behavior and psychology of different players. This can lead to severe variations in fraudulent and anti-fraudulent strategies, jeopardizing the win rate against top players.
    \item \textbf{Dependence on human expert knowledge:} For compressing the game tree in CFR to reduce computational cost, human expert knowledge on mathematics and computational theory is heavily needed, causing large knowledge bottleneck and labor effort for improvement.
\end{itemize}

Hence, there arises a pressing need for a lightweight model capable of autonomously extracting information to optimize IIGs further. Such a model should exhibit superior generalization performance, reduced dependence on expert knowledge, and enhanced adaptability to multi-player environments without compromising computational efficiency. Addressing this challenge will be a pivotal focus in future research within the field of Poker AI.


Thanks to the strong capability of large language model (LLM), it is promising to achieve a prominent framework for playing Poker in real games. LLM is a novel deep learning (DL) model that can handle massive amounts of natural language data, with various advantages in human-computer interaction. Firstly, by pre-training on large-scale text data, LLM can extract word vectors while also learning contextual information to enhance the natural language understanding capabilities. Secondly, LLM integrates generative tasks into the encoder-decoder structure to achieve a unified natural language processing capability, thereby improving the quality of text generation. Meanwhile, LLM exhibits flexibility and adaptability of human-computer interaction due to its remarkable zero-shot and few-shot abilities, that is, in-context learning that only changes the input rather than the parameters. Last but not least, end-to-end learning streamlines the need for intricate feature engineering and intermediate steps. LLMs can autonomously acquire task-relevant rules and properties leveraging the powerful representation capabilities of deep neural networks. Thus, it is feasible to develop an LLM-based approach for playing real games of multi-player no-limit Poker.

\begin{figure}[hbpt]
    \centering
    \includegraphics[width=1\linewidth]{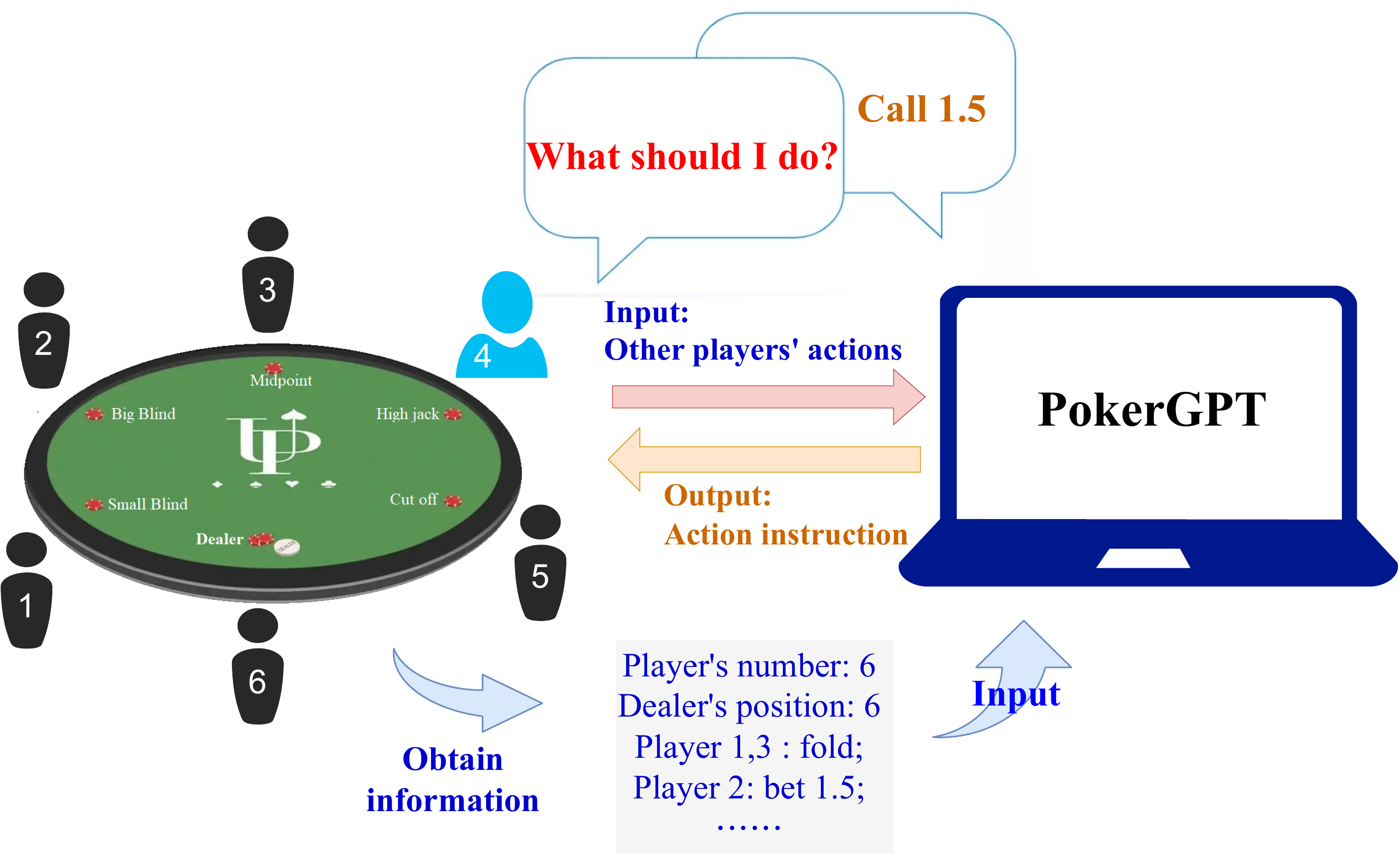}
    \caption{Interaction process of PokerGPT.}
    \label{interaction}
\end{figure}

In this paper, we utilize a lightweight LLM and data from real Poker games to construct an AI solver for Poker solutions, named PokerGPT. We firstly collect textual records of real games with arbitrary amount of players. Then, we filter necessary information and process these records into prompts, a textual data form that is understandable for both humans and LLMs. For dataset preparation, we integrate these prompts with actions in real games as labels. Based on the dataset, we fine-tune OPT-1.3B, a lightweight open-source LLM from Facebook~\cite{zhang2022opt} using reinforcement learning from human feedback (RLHF), a popular technique for fine-tuning LLMs. The interaction process is shown in Fig.~\ref{interaction}. Our codes are publicly available~\footnote{\url{https://github.com/hch211/TH_LLM}}.

The pivotal contributions of our work are listed as follows:
\begin{itemize}
    \item We present a novel approach based on LLMs to address IIGs. This approach employs an end-to-end learning method, proved to be easily trainable and considerably cost-effective compared with previous approaches.
    \item We propose PokerGPT, a pioneering DL model specifically designed for Poker. With the power of LLMs, our model has a prominent advantage on convenient interaction with humans. More importantly, PokerGPT has the capability of dealing with an arbitrary number of players in Poker games, showing the outstanding flexibility which has never been achieved by previous works. Besides, our model contains much less parameters and takes much less inference time compared with conventional methods such as CFR, while consumes much less training time than novel DL works, achieving great resource conservation.
    \item Based on data of real games, we devise an efficient data processing approach specific to Poker, consisting of data cleansing and prompt engineering, leading to significant improvements on fine-tuning of PokerGPT.
    \item Through experiments, we demonstrate that our model exhibits significant advantages over existing models in terms of various metrics, such as win rate, training speed, response time, etc. Furthermore, we design 2 metrics, named \textit{action score} and \textit{average investment}, to study characteristics of PokerGPT's strategies on Poker. 
\end{itemize}




The remainder of this paper is organised as follows: Section~\ref{related_work} gives a literature review of classical works of Poker and LLM; Section~\ref{prerequisites} introduces the rules of Poker and the principles of LLMs; Section~\ref{method} explains the holistic architecture of our proposed PokerGPT, including data acquisition, prompt engineering, and training process; Section~\ref{experiment} describes the case studies, shows performance of PokerGPT with multiple metrics, and analyzes the results; Section~\ref{conclusion} concludes this work and presents pathways for future work.

\section{Related Work}\label{related_work}
\subsection{Solutions for Poker}
Poker is a popular and challenging form of poker that has been used as a testbed for IIG algorithms for decades~\cite{rubin2011computer}. A major breakthrough in this field was the introduction of counterfactual regret minimization (CFR)~\cite{zinkevich2007regret}, a simple and efficient iterative algorithm that converges to a Nash equilibrium by minimizing the regrets of both players over time. CFR enabled the solution of heads-up limit Poker in 2015~\cite{bowling2015heads}, shifting the research focus to the more complex and realistic variant of Poker.

CFR once became a widely used algorithm for solving Poker, and a challenging benchmark for IIGs. Previous works have shown that CFR variants can achieve superhuman performance in Poker against professional human players. The work of~\cite{brown2017safe} introduced subgame-solving techniques that outperform prior methods both in theory and practice. By fusing these techniques, in~\cite{brown2018superhuman}, the authors presented Libratus, an AI based on Monte-Carlo CFR (MCCFR) that defeated 4 top human specialists in heads-up no-limit Poker in a 120000-hand competition. Though Libratus was the leading benchmark in IIG solving, it required millions of core hours and terabytes (TBs) of memory to calculate. To improve computational efficiency, the work of~\cite{brown2018depth} substituted values at leaf nodes of unknown states to limit the depth of the decision tree, and also outperformed prior methods. The work of~\cite{brown2019solving} introduced novel CFR variants to discount regrets from earlier iterations in some cases differently for positive and negative regrets, and reweight iterations to obtain the output strategies. Since games can be significantly complex when there are multiple players in games, the work of~\cite{brown2019superhuman} extended Libratus to Pluribus, an AI that defeated top human professionals in 6-player Poker, the most popular form of Poker. However, Pluribus approaches are also computationally expensive and memory-intensive, and consumed significant time for inference. 

To overcome the limitations of CFR-based algorithms, DL methods have been applied to exploit the power of big data and neural networks (NNs). For example, DeepStack~\cite{moravvcik2017deepstack} used an NN to approximate the CFR values and performed recursive reasoning, while deep CFR~\cite{brown2019deep} improved the performance and efficiency by using NNs to directly learn the CFR strategy in the full game without abstraction. To reduce the dependence on prior knowledge and training time, self-play deep reinforcement learning (DRL) was integrated with CFR to enhance the exploration~\cite{brown2020combining}. Moreover, the works of~\cite{zhao2022alphaholdem} proposed an end-to-end framework based on DRL and introduced several techniques to accelerate it and defeat DeepStack.

In conclusion, the above methods rely heavily on computational resources, prior knowledge, and domain-specific heuristics, which constrain the applicability. Besides, it is difficult for humans to use these works, also harming the practicality. Therefore, developing more efficient algorithms for Poker and other complex IIGs remains an open problem.


\subsection{LLM}
Recent advances in pre-trained language models, such as ChatGPT and GPT-4~\cite{openai2023gpt4}, have inspired a range of studies that leverage information from various domains and modalities. These studies aim to enhance the performance and applicability of language models for diverse tasks, and can be broadly divided into system design and ene-to-end training. System design employs ChatGPT as a dispatcher that orchestrates different expert models for specific tasks, such as Cola~\cite{chen2023language} and X-GPT~\cite{zou2023generalized}. On the other hand, end-to-end training integrates models from different modalities into a unified framework that can be trained end-to-end, and then fine-tuned on data from particular domains, such as weather forecasting~\cite{bi2023accurate} and augmented reality~\cite{li2023otter}.

The use of LLMs for games has attracted increasing attention from researchers, who have obtained some intriguing findings. For instance, the results of~\cite{srivastava2022beyond} showed that LLMs could produce responses that were remarkably similar to human feedback, despite some limitations on specific tasks in the Beyond the Imitation Game benchmark. This suggested that LLMs had the potential to handle human-related scenarios. The authors of~\cite{vartinen2022generating} leveraged GPT-2 and GPT-3 to procedurally generate RPG video game quest descriptions. Similarly, the authors of~\cite{kelly2023towards} used GPT-3 to provide dialogue suggestions for non-player characters as well as to highlight relevant game module information. They both found that GPT-3 could generate significantly better sentences than GPT-2, demonstrating the superior performance of GPT-3 and the promising prospects of future GPTs. Moreover, the authors of~\cite{akata2023playing} compared GPT-3, GPT-3.5, and GPT-4 models in playing finitely repeated games with each other and with other human-like strategies. They observed that LLMs performed sub-optimally in games that required coordination, but excelled in games that involved individual competition and self-interest, such as the iterated Prisoner’s Dilemma family. The above research implies that LLMs can be feasible as a Poker solver.

Although there are still many challenges regarding the use of LLMs for games, LLMs have shown great potential for game-related research and applications. In this work, we propose a framework for Poker via LLMs, emphasizing the potential for LLMs to solve IIGs

\section{Prerequisites}\label{prerequisites}
\subsection{Poker Rules}

\begin{figure}[hbpt]
    \centering
    \includegraphics[width=0.95\linewidth]{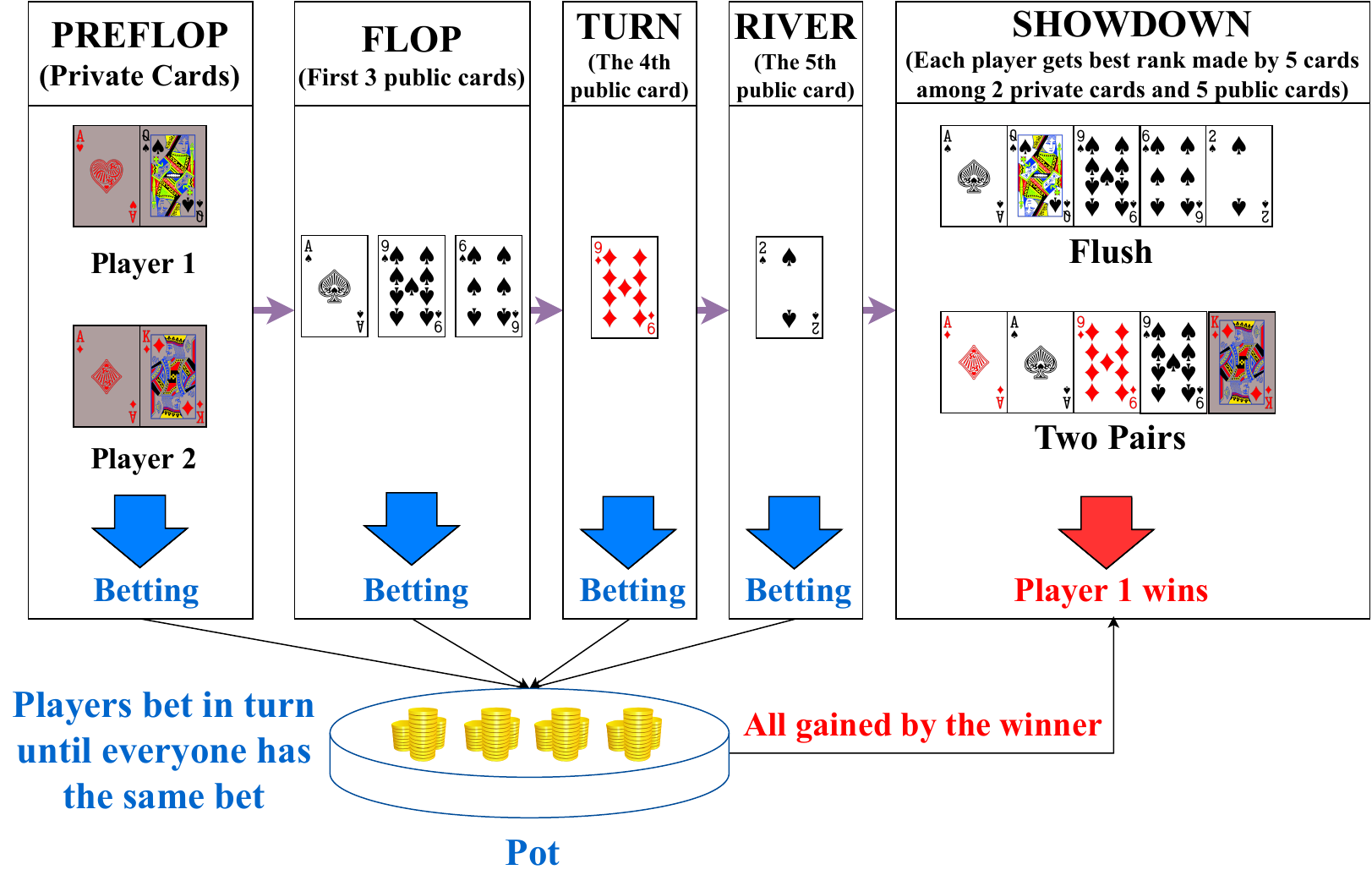}
    \caption{An example of a 2-player Poker game.}
    \label{example game}
\end{figure}

\begin{figure}[hbpt]
    \centering
    \includegraphics[width=0.95\linewidth]{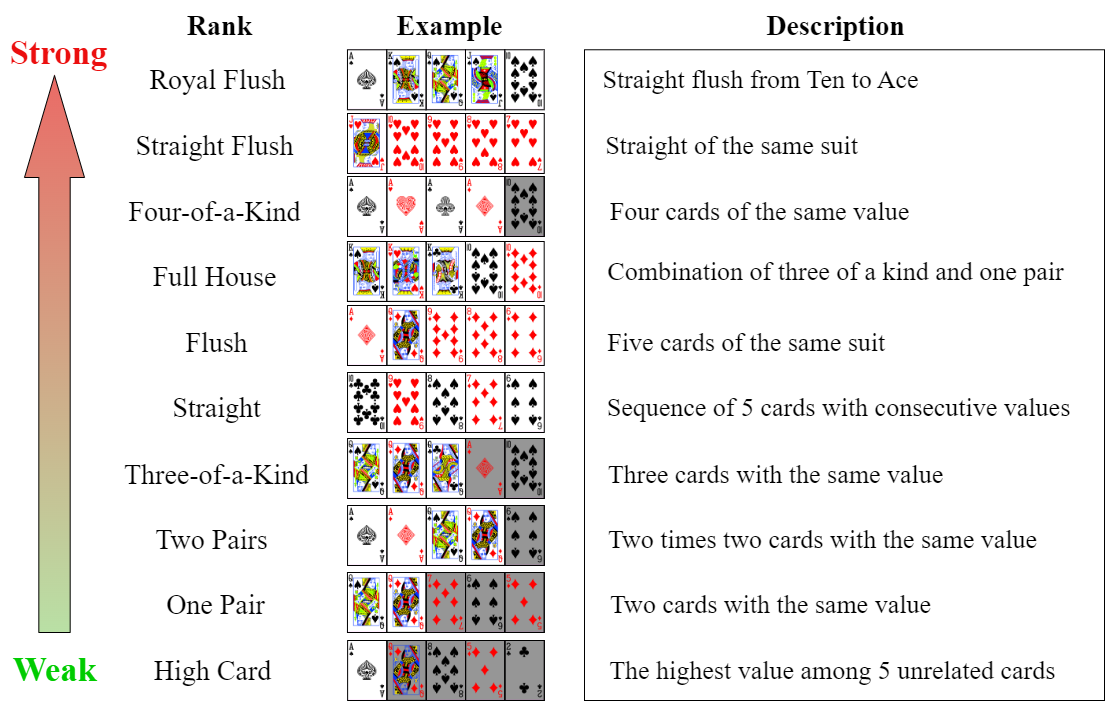}
    \caption{\vspace{8pt} Poker card strengths. \vspace{-10pt}}
    \label{card strength}
\end{figure}

Poker is a game that consists of multiple rounds, each starting with 2 private cards (``PREFLOP'' cards) dealt to each player, followed by 5 public cards (community cards) dealt in 3 stages. The stages are a set of 3 cards (the ``FLOP''), an additional single card (the ``TURN''), and a final card (the ``RIVER''). Each player aims to form the best 5-card, also named as \textbf{a hand}, using 2 private cards and 5 public cards. Players can ``bet'', ``check'', ``call'', ``raise'', or ``fold'' at each round. The player who has the best hand at the ``SHOWDOWN'', or who remains in the game after all others ``fold'', wins the money in the pot. Fig.~\ref{example game} illustrates an simple example of a 2-player Poker game, and the bet size is limited only by the total amount wagered in each round. Besides, Fig.~\ref{card strength} shows the card strengths.

\subsection{LLM}
LLM is a kind of DNN that generates natural language texts from various inputs, such as prompts, queries, or contexts. LLMs learn from massive text corpora that cover diverse domains, such as books, news, articles, web pages, and social media posts. LLMs can perform various natural language processing tasks, such as text generation, text summarization, question answering, and machine translation. The main process of training a LLM is data preparation, pre-training, and fine-tuning.

\subsubsection{Pre-training}
LLMs are pre-trained with self-supervised learning, which leverages the data itself as supervision. Self-supervised learning can capture different levels of language representation, such as words, sentences, or documents. Common self-supervised learning objectives for LLMs include masked language modeling, next sentence prediction, permutation language modeling, causal language modeling, etc. Since our work focuses on fine-tuning and RLHF, we will not introduce pre-training of LLM in details.

\subsubsection{Fine-tuning}
LLMs are fine-tuned with supervised or semi-supervised learning, which leverages labeled or partially labeled data as supervision. Supervised or semi-supervised learning can adapt LLMs to different natural language processing tasks. After fine-tuning, LLMs can be used for inference. Common inference tasks include text summarization, question answering, machine translation, text generation, etc.

\subsubsection{RLHF}
A popular ML technique for LLM is RLHF, that trains a ``reward model'' based on human feedback as a reward function to optimize an agent's policy using reinforcement learning (RL) through an optimization algorithm~\cite{bai2022training,abramson2022improving}. This method can improve the robustness and exploration of RL agents, especially when the reward function is sparse or noisy. It is used in tasks where it is difficult to define a clear, algorithmic solution but where humans can easily judge the quality of the model's output.

\begin{figure*}[hbpt]
    \centering
    \includegraphics[width=1\linewidth]{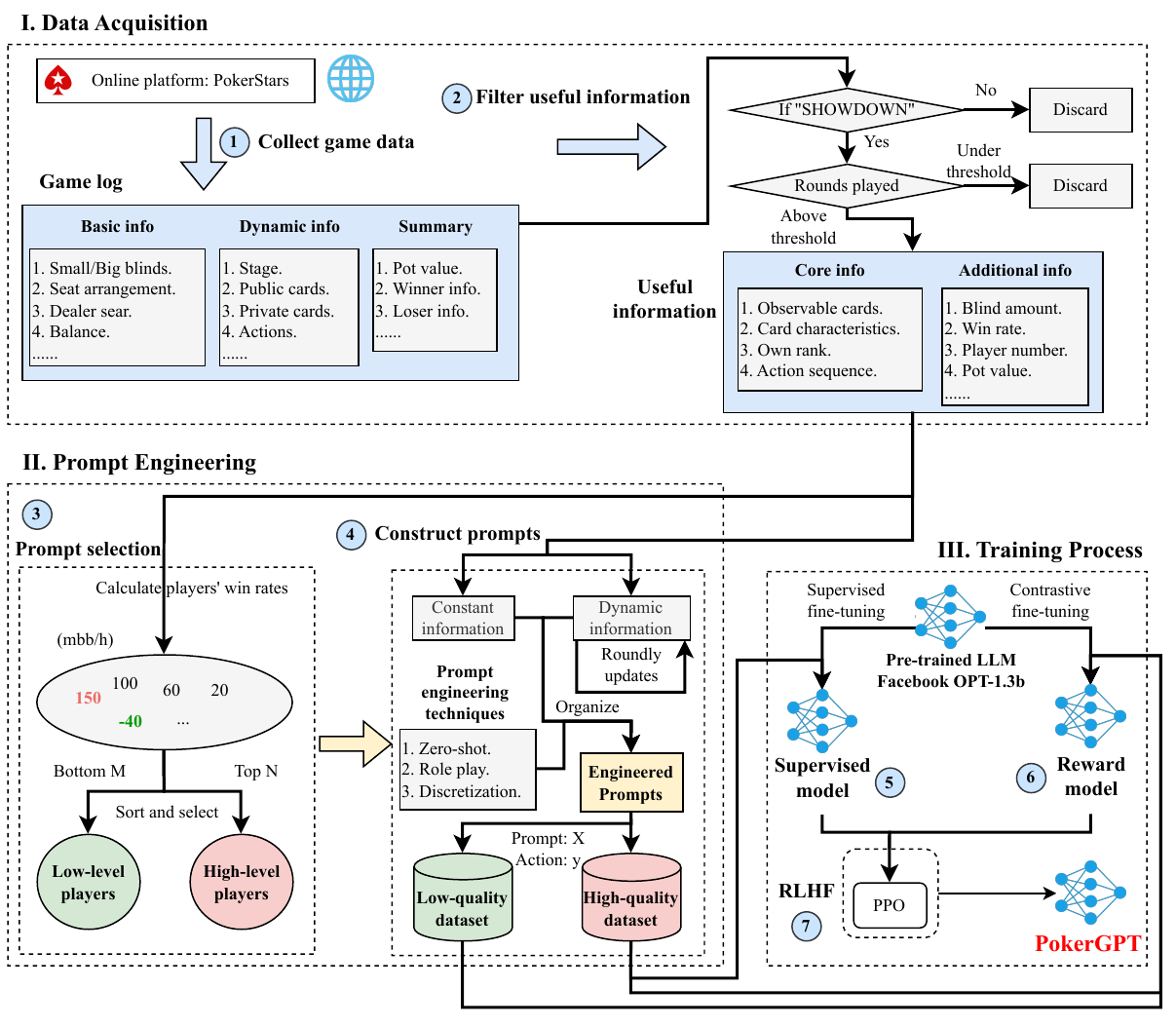}
    \captionsetup{width=.8\linewidth}
    \caption{The framework of PokerGPT, consisting of data acquisition, prompt engineering, and training. \vspace{-10pt}}
    \label{Fig:method}
\end{figure*}

\section{PokerGPT}\label{method}
In this section, we introduce our framework, PokerGPT, in details. We firstly collect a group of game data on a online Poker platform in real world~\footnote[2]{PokerStars: \url{https://www.pokerstars.com}}. Then we transform the data into prompts which can be understood by both humans and LLMs. To improve performance, we select prompts with high quality for further training. By utilizing supervised fine-tuning, we make a pre-trained LLM capable of making decisions when receiving staged information provided by players in a Poker game. For generalizability and interactivity, we adopt RLHF to enhance the supervised model. Besides, since the LLM requires as much information as possible while the interaction may frequently happen in a short period of time, we set all the information as fixed instruction for simplicity. By doing this, humans can interact with our model through just simple dialogues. Finally, we conduct supervised fine-tuning, reward modeling, and RLHF to get PokerGPT. The framework of PokerGPT is demonstrated in Fig.~\ref{Fig:method}.

\subsection{Data Acquisition}

\subsubsection{Data Description}
For increasing the practicality of our model, we acquire logs of Poker game on PokerStars. These logs contain information as follows:
\begin{itemize}
    \item \textbf{Basic information}: This encompasses the establishment of both the big and small blinds, the seating arrangement details for each player, along with their respective chip counts and the prevailing position of the current dealer. This segment furnishes a thorough comprehension of the entire gaming scenario.
    \item \textbf{Dynamic information}: For each game, we partition it into distinct stages. Initially, there is the ``PREFLOP'' stage, during which each player receives 2 private cards, initiating the first round of betting. Subsequently, the ``FLOP'', ``TURN'', and ``RIVER'' stages unfold, wherein the community cards are progressively revealed, accompanied by the betting actions of the remaining players. The conclusive stage ``SHOWDOWN'' unveils the private cards, ultimately determining the definitive winner, without any action should be done by players.
    \item \textbf{Summary}: An overview of the game pool size, the community cards, and each player's earnings is recorded. It includes details on the winnings and losses associated with specific card combinations, instances of non-participation, and other relevant information.
\end{itemize}

\begin{figure}[hbpt]
    \centering
    \includegraphics[width=1\linewidth]{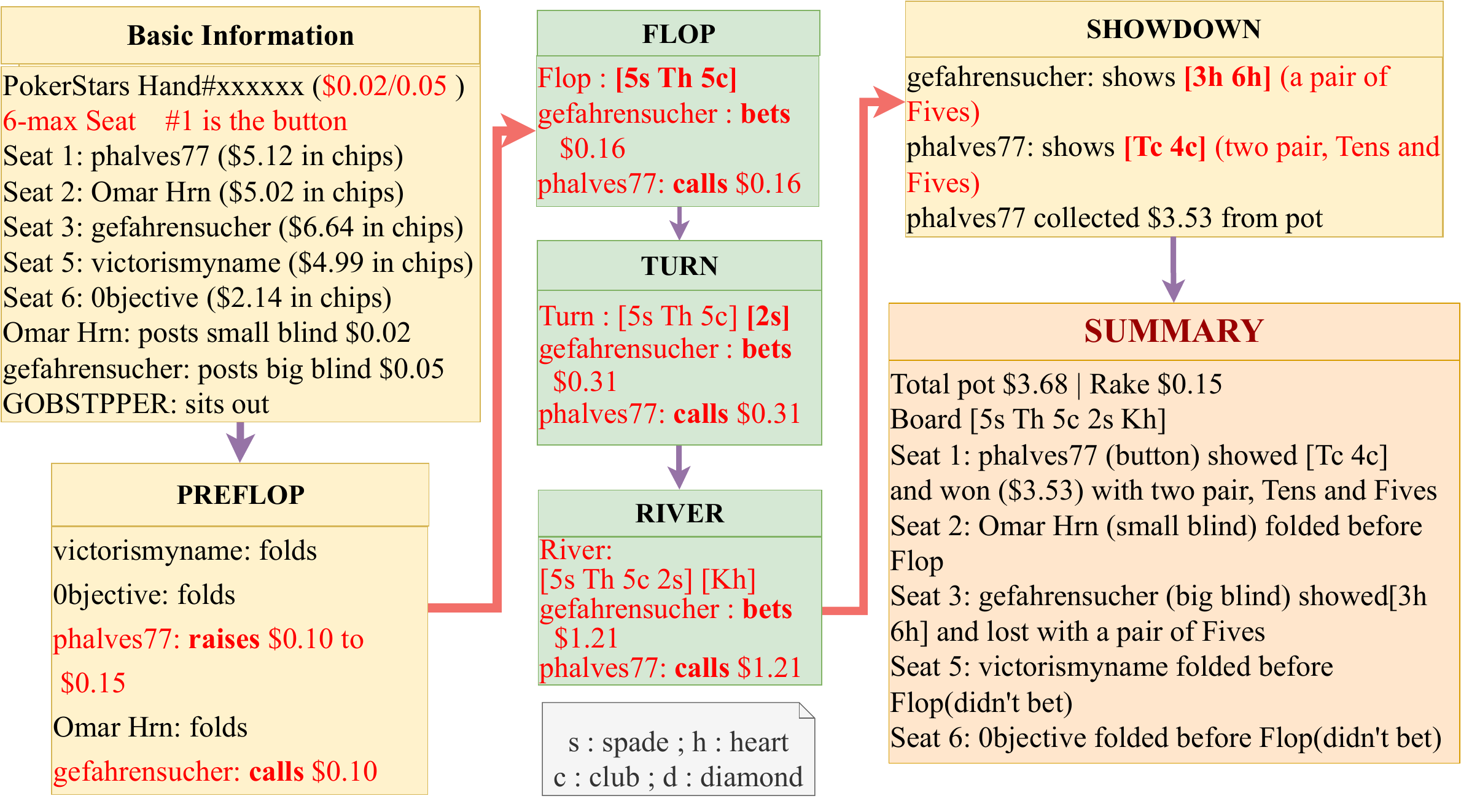}
    \caption{An example of Poker game log. \vspace{-10pt}}
    \label{game log}
\end{figure}

Fig.\ref{game log} shows an example of the game log. The chart reveals that the blinds for this game are set as \$0.02/\$0.05, with the dealer positioned at seat number 1. The chip count for each player is specified. For instance, the player in seat 1, \textit{phalves77}, holds 5.12 in chips.

Details on players' decisions and actions are as follows: During the ``PREFLOP'' stage, \textit{phalves77}, seated in the dealer's position, opted to ``raise'' to 0.15, while \textit{gefahrensucher} spent another 0.10 to ``call'' from the big blind. The community cards unfolded as follows: the ``FLOP'' displayed [`5s', `Th', `5c'], followed by the ``TURN'' card [`2s'], and finally the ``RIVER'' card [`Kh']. After each round of public cards, the recorded information includes the amount of each player's bet, and their action choices. For instance, after the ``FLOP'', \textit{gefahrensucher} give the ``bet'' of 0.16, with \textit{phalves77} deciding to ``call''. After the ``TURN'' and ``RIVER'', \textit{gefahrensucher} placed the ``bet'' of 0.31 and 1.21, respectively, with \textit{phalves77} choosing ``call'' each time. During the ``SHOWDOWN'', \textit{gefahrensucher} revealed a pair of `5's (Fives), while \textit{phalves77} presented 2 pairs: `10's (Tens) and Fives. Consequently, \textit{phalves77} claimed the pot's bottom as the winner, gaining victory with the superior 2 pairs.

\subsubsection{Information Filtering}\label{information filtering}
The essential information of Poker games revolves around the public cards and the behaviors of the players in each game. The public cards stand as the pivotal information, forming an integral component of a player's ultimate hand. In Poker games, the unveiling of public cards directly influences the rank of each player's hand. The actions of opponents during each round (``bet'', ``raise'', ``check'', ``call'', ``fold'') serve as vital indicators to infer their strategies and intentions, directly providing insights into the strength of their cards. Moreover, the magnitude of money placed exhibits a player's confidence and card strength in a large extent. Also, the frequency of ``raise'' can enhance the understanding of opponent's playing style and card strength, enabling strategic decisions on when to bluff and when to discern a bluff.

\textbf{Core information} of utmost significance is crucial to be included in the prompt, listed as follows: 
\begin{itemize}
    \item \textbf{Observable cards}: It is necessary to know the 2 private cards and public cards exhibited. 
    \item \textbf{The characteristics of cards}: We describe the characteristics of private cards in several ways. ``Suit'' means 2 private cards with the same color. ``High'' means possessing one card higher than `9'. ``Close" means the difference of 2 cards is less than 5, which can form a ``straight''. 
    \item \textbf{The rank of the hand}: Since it may be difficult for the LLM to learn to recognize the rank of the hand, we simply provide the rank in the prompt. The rank can guide LLMs to understand the strength of different combinations of private cards and public cards. 
    \item \textbf{Action sequence}: Analyzing players' behavioral patterns plays a pivotal role in understanding the psychology and strategy of opponents. Thus, it is crucial to record each player's sequential actions.
\end{itemize}

\textbf{Additional information} also assists the learning and understanding of our model, including:
\begin{itemize}
    \item \textbf{Amounts of blinds}: Small and big blinds is important for evaluating both the revenue and the risk of Poker games, thereby contributing to the understanding of the decisions on ``raise'' or ``fold''.
    \item \textbf{Positions of players}: The positional context of a player relative to the dealer, coupled with the player's actions, serves as an indicator of the hand's strength, which influences the players' advantages and the game results.
    \item \textbf{Number of players}: The variation in the number of players within a game intensely changes the game structure. It is essential to make our model aware of and respond to the variation.
    \item \textbf{Balance}: We record each player's balance through the initial amount and the amount each has invested in the pot.
    \item \textbf{Discard}: The ``fold'' chosen by players helps our model to determine the remaining players.
    \item \textbf{Pot value}: The total amount of money put into the pot helps LLM understand the concept of odds.
    \item \textbf{Stage}: The same action in different stages generally has different purposes.
\end{itemize}


The incorporation of the above information ensures an effective training process for our model, improving language generation and comprehension capabilities.

\subsection{Prompt Engineering}
One of the advantages of LLMs over traditional DL models is that it has good interactivity, which can be achieved by processing verbose raw data into prompts and further used for fine-tuning. A prompt is a corpus that is understandable by both humans and LLMs, which is easy to be produced, such as colloquial phrases, sentences, and paragraphs. Therefore, we can quickly get suggestions on playing Poker games by inputting simple words with part of important information of a Poker game into LLMs. We will introduce how we make raw data into prompts, and filter out high-quality prompts to ensure that our model is able to give high-revenue or low-loss insights in real-time Poker games.

\subsubsection{Prompt Selection}

Based on the game log presented in Fig.~\ref{game log}, we can observe that the information contained in each game is diversified, and it is difficult for a pre-trained LLM to distinguish which information to focus on. Thus, we construct prompts with rich information and then select high-quality ones for training.

Firstly, we conduct searching for all logs to filter out ``SHOWDOWN'' stages that with private cards publicated. Cases without a specific winning or losing hand were subsequently excluded, because the lack of hand strength information makes it difficult for LLMs to learn optimal strategies in an observable situation.


Then, to get high-quality dataset, we adopt win rate for selection of high-level players. Win rate is a standard metric in the Poker AI community. We firstly counted the total number of games played by each player in the obtained logs. Next, we calculate overall revenue and converting it to a multiple of big blinds. After that, we count each player's win rate in milli-big-blinds per hand (mbb/h) as a sorting criterion for descending rankings. To mitigate bias resulting from a limited number of games, we exclude players whose number of games is under a threshold. Then, we select a group of top players as our high-quality dataset, in which the winning games have significant instructive impacts on our model. Note that, it is fine for our model to learn player actions in lost games. Intuitively, the way of high-level player losing games also has reference value. To some extent, it can help the model learn how to lose less.

On the other hand, we also collect data of the worst players as negative examples. Through reward modeling which will be mentioned in Section.~\ref{RLHF}, the actions and ideas of the low-level players can be learned and avoided by our model.

\subsubsection{Prompt Construction}



We choose zero-shot learning for prompt construction, which simply feeds the text into the model and ask for results. For practicality, we discard few-shot prompting. The purpose of presenting few-shot examples in the prompt is to describe the task instruction to the model in the form of demonstrations. Though adding question-answering (QA) samples to prompts may greatly improve the response quality of LLMs~\cite{zhao2021calibrate}, in real Poker games, the player's decision-making time is generally about 15 seconds, causing that players are unlikely to be able to input a large amount of QA samples or game information. Thus, for enhancement, we just add fixed instructions in prompts to explain our intent to the model. During interaction, prompts are systematically constructed mainly using the information in red in in Fig.~\ref{game log}. Besides, few-shot learning can be expensive in terms of token usage and violate the length limitation, while rich-information instruction improves the model to be more aligned with human intention and greatly reduces the cost of communication. Thus, directly giving instructions is a promising approach. On the other hand, due to the fine-tuning we conduct, explained in Section~\ref{RLHF}, few-shot learning is not necessary.

In terms of reducing tokens inputted, we roughly divide useful information into constant information which is fixed at the beginning of one game, and dynamic information updated in each decision round.

Furthermore, multiple prompt engineering techniques adopted are outlined as follows, which are proved to be effective in LLM training \cite{lilianweng}:
\begin{itemize}
    \item \textbf{Role play}: We give a role to our model and add corresponding text into our prompt, such as ``You are a professional gambler".
    \item \textbf{Value discretization}: To reduce the learning difficulty, we discretize the money needed to be determined by our model through rounding up the values of money paid in each stage to an integer multiple of the big blind, transforming the regression task into a classification task.
\end{itemize}

\begin{figure}[htbp]
\centering
\resizebox{\linewidth}{!}{
    \begin{AIbox}{Prompt Instance}
    \parbox[t]{\linewidth}{
    You are an experienced gambler. Now you need to assist me to make decisions in Texas Hold'em games. You have been provided with a series of observable information:\\
    \small{Player amount: [6], Currency: USD, Blind value: [0.02/0.05], Order: [`2', `3', `5', `6', `7', `9'], Seat 2 is small blind.} \\
    \small{My cards: [`Th', `Ah'], the characteristics of my cards: [``suit'', ``high'', ``close''], My seat: [Seat 2]}\\
    \small{Stage: ``PREFLOP'', Public cards: [`**' `**' `**' `**' `**']}\\
    \small{My rank: [``High''], Money: [3.92], Action: []}\\
    \small{Seat 3: [`**', `**'], Money: [2.33], Action: [], Discard: [False]}\\
    \small{Seat 5: [`**', `**'], Money: [5.54], Action: [], Discard: [False]}\\
    \small{Seat 6: [`**', `**'], Money: [3.75], Action: [], Discard: [False]}\\
    \small{Seat 7: [`**', `**'], Money: [4.22], Action: [], Discard: [False]}\\
    \small{Seat 9: [`**', `**'], Money: [1.47], Action: [``raises 0.05 to 0.1''], Discard: [False]}\\    
    The pot value is [0.17]\\
    The actions can be: [``fold'', ``raise'', ``call'']. What should I do? If I choose to ``bet'' or ``raise'', then how much? Choose a number from \{(0, 0.05, 0.15, 0.3, 0.5, 1, 2.5, \textit{3.92}\}.
    }
    \end{AIbox}}
    \captionsetup{width=.99\linewidth}
    \caption{An instance of our prompt. \vspace{-10pt}}
    \label{fig:prompt}
    \end{figure}

An textual instance of our prompt is shown in Fig.~\ref{fig:prompt}. It is worth noting that in real games, sometimes opponents may directly show 1 or 2 privates cards consider from some psychological warfare perspectives. In that case, we can replace the ``**'' with the displayed cards. An utilization example of the engineered prompt can be found in Section~\ref{experiment}, shown as Fig.~\ref{fig:PokerGPT}. 

The information received by our model encompasses all core information and additional information mentioned in Section~\ref{information filtering}. This array of data enables our model to comprehensively understand the prevailing situation.



\subsection{Training Process}\label{RLHF}
We utilize the prepared dataset to train our model using DeepSpeed-Chat~\cite{yao2023deepspeed}. DeepSpeed-Chat is a system framework that enables an end-to-end training experience for ChatGPT-like models. The main steps consist of: (1) Supervised fine-tune a pre-trained LLM on small datasets to provide specific prior knowledge to this model; (2) Train a reward model using data separated by quality. (3) Apply RLHF to make the fine-tuned LLM in step (1) able to understand various human text assisted by the reward model in step (2). The pre-trained model we choose is the OPT-1.3B from Facebook, which performs similarly to GPT-3\cite{zhang2022opt}, making the goal of training domain-specific expert-level AI assistant possible. The reason for choosing this model is its small size and good comprehension in human dialogues.

Supervised fine-tuning is a necessary approach because re-training a LLM which has been already pre-trained consumes a substantial amount of computational resources. On the other hand, involving domain-specific expert model training necessitates not large-quantity but high-quality data that contains comprehensive details of the designated task. Particularly in the supervised fine-tuning phase, a modest number ranging from tens to hundreds of high-quality data proves sufficient for achieving satisfactory results in model refinement~\cite{openai2023gpt4}. Thus, it is promising to fine-tune OPT-1.3B on our collected dataset.

The second step is to train a reward model, which is crucial for the RLHF process. We need to train a model to predict the reward signal, which is used to guide the fine-tuning of the final dialogue model. For convenience, we directly use the win rates calculated before as the labels of our reward model. Then, we input prompt-action pairs into the reward model to generate a score indicating which action is better. The model is also trained using supervised learning.

Finally, RLHF fine-tuning is applied to train the supervised fine-tuned model based on the reward model in the second step. Specifically, fine-tuning approach used here is proximal policy optimization (PPO)~\cite{schulman2017proximal}, a popular and strong RL algorithm, whose reward is generated by the reward model. For updating, RLHF involves generating new answers with the current model, ranking these answers using the reward model, and then updating the model's parameters to favor answers with higher reward. This iterative process improves our model from the perspective of robustness to various human-like input, and generalizability to extensive game scale and varying situations. In the end, we obtain the PokerGPT for producing human-like and engaging answers to instruct Poker games.

\section{Experiment}\label{experiment}
\subsection{Fundamental Setup}
We firstly prepare our dataset based on data of over 1 million games. After processing mentioned in Section~\ref{method}, we get high quality data of about 120 thousand games. Then, we divide 90\% of them into training set and the remaining 10\% into test set. We select Microsoft's DeepSpeed-chat framework~\cite{yao2023deepspeed} for its comprehensive functionalities. This framework also efficiently manages the distribution of model parameters across individual GPUs during training. We fine-tune an open-source LLM 1.3 billion parameters on HuggingFace, named Facebook's OPT-1.3B model, known for its comparable performance to GPT-3~\cite{zhang2022opt}. We run our experiments on a single NVIDIA GEFORCE 3090. The hyperparameter we use can be refered to DeepSpeed-Chat~\cite{yao2023deepspeed}, while we change mini batch size to 4 due to the limitation of our computational device.

\subsection{Experimental Setup}

\subsubsection{Performance Comparison}
For performance comparison, we make PokerGPT play against Slumbot~\cite{jackson2013slumbot} for 10000 hands, and compare its results with 2 latest models~\cite{brown2020combining,zhao2022alphaholdem}. 

\subsubsection{Ablation Studies}
We aim to study if useful information filtering and prompt engineering can improve the performance of PokerGPT. Thus, we input raw game data and filtered data into LLMs for fine-tuning, and compare the win rate. Furthermore, as mentioned in Section~\ref{method}, we use win rate for evaluating the level of players included in our dataset. To explore if levels of players have impact on the fine-tuning results, we sample data from 50000 games based on different win rate thresholds to construct our training set for fine-tuning. All datasets for fine-tuning are demonstrated in Table~\ref{Table_dataset}. As for test set, we use Slumbot for win rate and randomly sample 1000 games of winners among the complete dataset for other metrics introduced as follows.

\begin{table}[hbpt]
\centering
\caption{Datasets under different setup for fine-tuning.}
\setlength{\tabcolsep}{3mm}{
\renewcommand\arraystretch{1.2}
\begin{tabular}{cccc}
\hline\hline
\textbf{Dataset} & \textbf{Technique} & \textbf{Win Rate Threshold} & \textbf{Amount}  \\ \hline
I   & Raw & - & 1 million         \\
II & Info filtering & - & 1 million \\
III  & Win rate sorting & $>1500$ mbb/h & 50000        \\
IV  & Win rate sorting &  $600 \sim 1200$ mbb/h & 50000  \\
V & Win rate sorting &  $0 \sim 500$ mbb/h & 50000 \\ 
VI & Win rate sorting  &  $<0$ mbb/h & 50000 \\ 
\hline\hline
\end{tabular}}
\label{Table_dataset}
\end{table}

\begin{table*}[htbp]
\centering
\caption{Performance comparison against the latest 2 models.}
\setlength{\tabcolsep}{1.5mm}{
\renewcommand\arraystretch{1.3}
\begin{tabular}{cccccccc}
\hline\hline
\textbf{Model}            & \textbf{Win Rate (mbb/h)}     & \textbf{Training Time (h)} & \textbf{Response Speed (s)} & \textbf{Player Amount}   & \textbf{Parameter Amount} & \textbf{Main Tech} & \textbf{Year} \\
\hline
ReBel~\cite{brown2020combining}                     & 45$\pm$5                  & -                      & -                       & 2                        & -                         & DRL+CFR            & 2020          \\
AlphaHoldem~\cite{zhao2022alphaholdem}               & 111$\pm$16                & 580                    & 0.017                   & 2                        & 8.6 million               & DRL                & 2022          \\
{\ul \textbf{PokerGPT}} & {\ul \textbf{158$\pm$49}} & {\ul \textbf{9.5}}     & 5.4                     & {\ul \textbf{2 or more}} & 1.3 billion               & LLM                & 2023   \\
\hline\hline
\end{tabular}}
\label{Table_Comparison}
\end{table*}

To evaluate the performance of supervised fine-tuning on datasets of different quality, we regard determining actions, including ``bet'', ``fold'', ``raise'', and ``check'', as a multi-label classification task. If our model gives a action different from the real action in test set, we get a ``False''. We take macro F1 as the metric for this task~\cite{yang1999evaluation}. The macro F1 score is a type of F1 score that gives equal weight to each class in multi-label classification, regardless of the class distribution. Assume we have $N$ classes, the macro F1 can be denoted as:
\begin{align}
    \text{macro-F1}=\frac{1}{N}\sum^N_{i=1}\text{F1}_i, \\
    \text{F1}_i=\frac{2 \times \text{precision} \times \text{recall}}{\text{precision}+\text{recall}},
\end{align}
where $F1_i$ is the F1 score of the $i$th class. Since we have 5 actions for Poker games, $N$ equals to 5 here. Note that, for 50000 games, each action which is regarded as classification label may have different amount of labels from 50000.

On the other hand, we study on value differences of value-related actions from all correct predictions, such as that our model says ``bet \$1.5'', while the real action in test set is ``bet \$2.0''. Since there exist multiple currencies and big blind values in our data, we unify currency to US dollar to represent all money using big blind. As mentioned before, we discretize values of ``bet'' and ``raise'' into multiple of big blind. Specifically, here we adopt \{0, 1, 3, 6, 10, 20, 50, 100, \text{all-in}\} as the multiples of big blind for decision, while the values exceeding user's owned money will be excluded. As we regard value prediction as a multi-label classification task, we compare the frequency of each multiple chosen to evaluate the accuracy of predicted values, as well as using mean squared error (MSE) for a general evaluation.

Besides, perplexity in Natural Language Processing (NLP) is a measure of how well a probability model predicts a sample. Assume we have a sequence $W$ with $N$ words, a perplexity is formulated as:
\begin{align}
    \text{perplexity}(W) = P(w_1,w_2,...,w_N)^{-\frac{1}{N}},
\end{align}
where $W=(w_1,w_2,...,w_N)$ is the sequence of words, and $P$ is the probability. It quantifies the uncertainty of a model by calculating the inverse probability of the test set, normalized by the number of words. In the context of language models, a lower perplexity generally indicates the model is better at predicting the next word in a sentence.

\subsubsection{Multi-Player Evaluation}
We evaluate PokerGPT in multi-player games using an open-source Poker simulation~\footnote[3]{\url{https://github.com/dickreuter/neuron_Poker}}. For setup consistency, all bots in multi-player experiments are based on a mandatory method that calculates equity. To study the strategy variation of PokerGPT during player amount changes, we design 2 metrics, action score and average investment, to represent the occurring frequency of each action and the money used in each game, relatively:
\begin{itemize}
    \item \textbf{Action score:} Since some actions can occur multiple times in one game, such as ``check'', ``raise'', ``bet'', and ``call'', while the others can only occur once, such as ``all-in'' and ``fold'', the length of one game should be considered. Thus, as a game can have up to 4 stages, we calculate action score for each action by dividing the number of times this action occurs in a game by the stage number of this game. For example, if a player do ``check'' for 3 times and ``fold'' in ``TURN'' stage, the action score for ``check'' is 1, and the action score for ``fold'' is 0.33. Finally, we divide action scores by the amount of games, to make action scores ranged from 0 to 1.
    \item \textbf{Average investment:} In Poker games, more money usage means both higher revenue and higher loss. By observing the amount of average investment, we can know the style of playing more clearly. To get the average investment, we calculate the mean of total money used by our model in each game.
\end{itemize}
By analyzing action scores and average investment, we can get more detailed insights of our model's strategies for different player amount.


\subsubsection{Analysis on Interaction}
We show a case of PokerGPT interaction with multiple players in processed dataset. The aim is to evaluate the general capability of our model through analyzing the answers of different questions.

\subsection{Comparison with Previous Works}

\begin{figure}[htbp]
    \centering
    \includegraphics[width=1\linewidth]{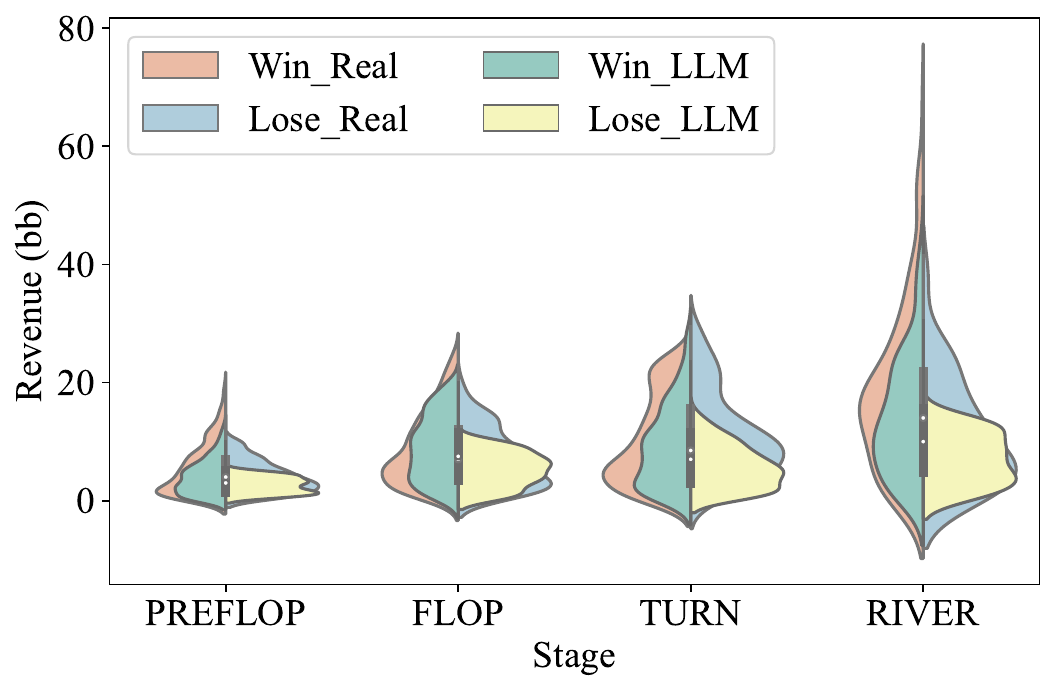}
    \captionsetup{width=.99\linewidth}
    \caption{The distribution of revenue among 1000 games with unit of big blind. The real results are counted from the collected dataset, while the results of PokerGPT are from the games against Slumbot.}
    \label{Fig:revenue_distribution}
\end{figure}

Firstly, we compare our performance against Slumbot with 2 novel works, recording multiple metrics in Table~\ref{Table_Comparison}. We can see that our model makes progress in win rate. Compared to Rebel (45+5) and Alphaholdem (111+16) against Slumbot, our model comes out on top with a win rate of 158+49, measured in mbb/h, showing the superiority of our model. However, since the standard deviation of our model is higher than the other 2 models, fluctuation of our performance is more violent, suggesting the larger variation of the LLM's answers led by the diversified nature of generative models, which should be further optimized. Furthermore, our model completes the training in just 9.5 GPU hours (3.5 for supervised fine-tuning, 1.5 for reward model training, and 4.5 for RLHF), compared to 580 GPU hours for Alphaholdem, showing the improvement on training efficiency. Thanks to the adaptability of LLMs, empowering LLMs with specific prior knowledge through fine-tuning on a relatively small scale of data is effective. Last but not least, our model is capable of providing action instructions for arbitrary number of players, while the other 2 models are limited to playing one-on-one situations. Even without transferring that is needed for the other 2 models, LLM-based model can be directly generalized to various situations, saving a lot of computational resources. On the other hand, though PokerGPT owns a much larger amount of parameters and a slower response speed than Alphaholdem, both metrics are acceptable in real games, and much better than previous CFR-based works. Additionally, we illustrate the staged revenue distributions of high-level players in real games and PokerGPT against Slumbot in Fig.~\ref{Fig:revenue_distribution}. Compared with human players, PokerGPT tends to keep a more stable win rate by losing less in later stages, while humans are more likely to take a risk in gambling in later stages, since the later the stage is, both the revenue and the risk can be higher.

\subsection{Ablation Results}

\begin{table}[htbp]
\centering
\caption{Ablation on different dataset setup.}
\setlength{\tabcolsep}{2.4mm}{
\renewcommand\arraystretch{1.2}
\begin{tabular}{ccccc}
\hline\hline
\textbf{Dataset} & \textbf{Win Rate (mbb/h)} & \textbf{Macro F1} & \textbf{MSE} & \textbf{Perplexity}\\ \hline
I                & -376$\pm$215          & 31.34             & 115.41  & 4.25 \\
II               & 229$\pm$137           & 62.77             & 36.77   & 2.71 \\
III              & 1221$\pm$54           & 77.63             & 0.21    & 1.03 \\
IV               & 1034$\pm$52           & 72.22             & 0.95    & 1.04 \\
V                & 608$\pm$59            & 68.15             & 8.28    & 1.05 \\
VI               & -574$\pm$76           & 59.69             & 177.26  & 1.09 \\
\hline\hline
\end{tabular}}
\label{Table_ablation}
\end{table}

\begin{figure}[hbpt]
    \centering   
    \subfigure {
    \includegraphics[width=.205\textwidth]{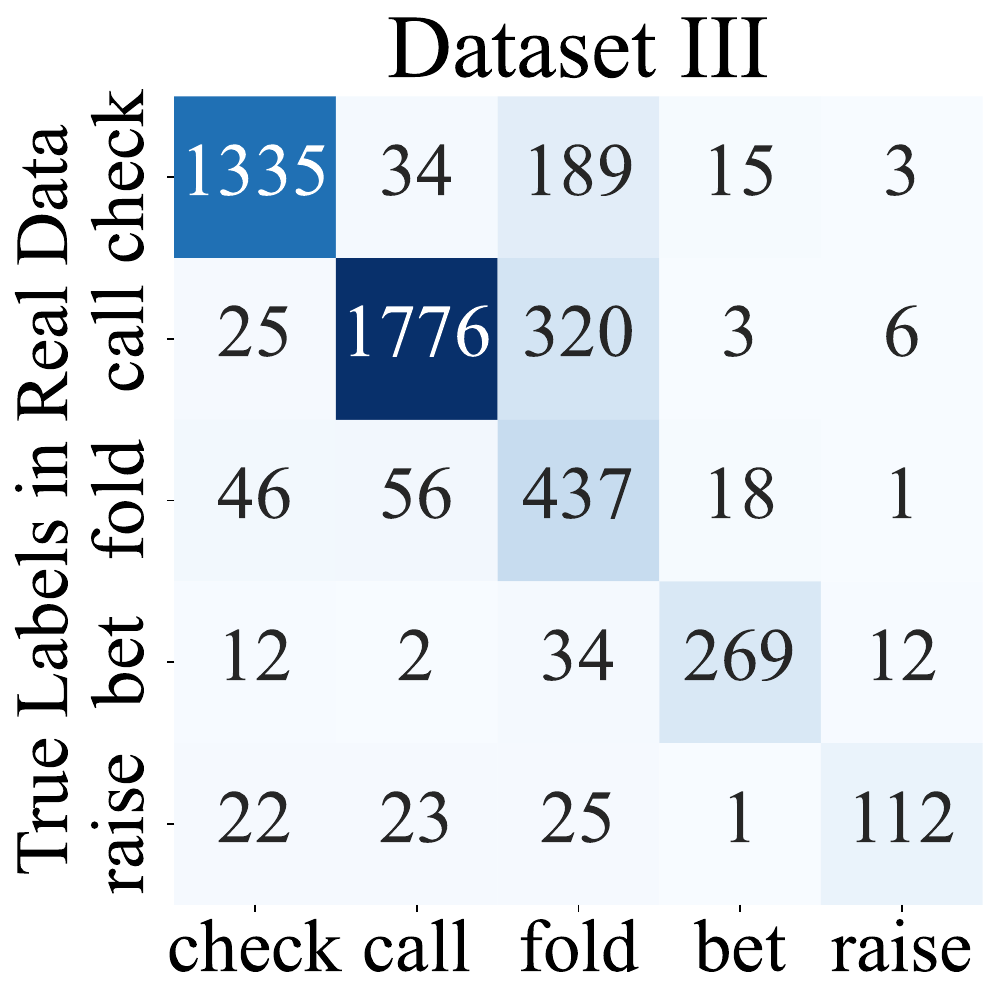}  }
    \hspace{-0.1in}
    \subfigure {
    \includegraphics[width=.24\textwidth]{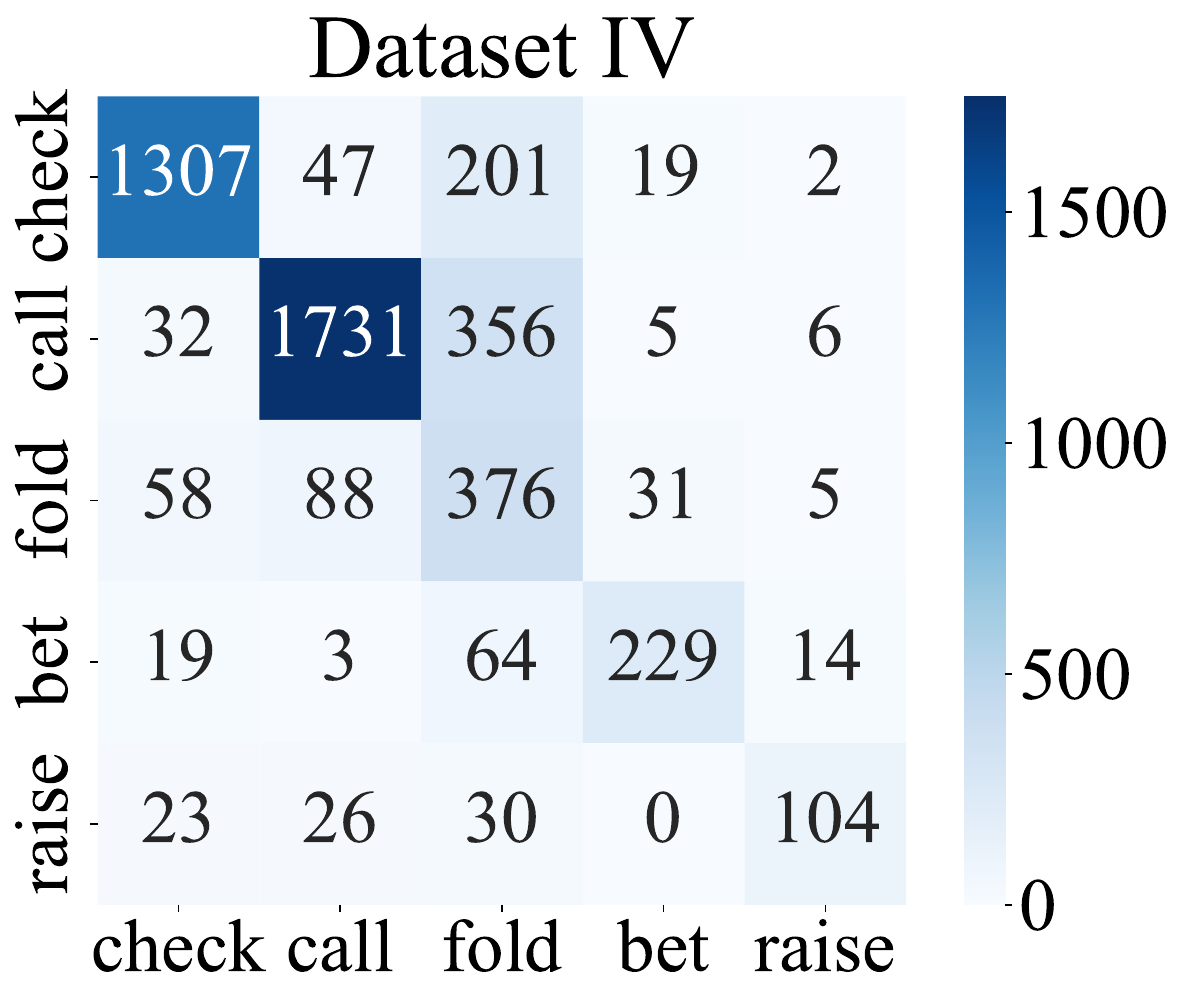}  }
    \subfigure {
    \includegraphics[width=.205\textwidth]{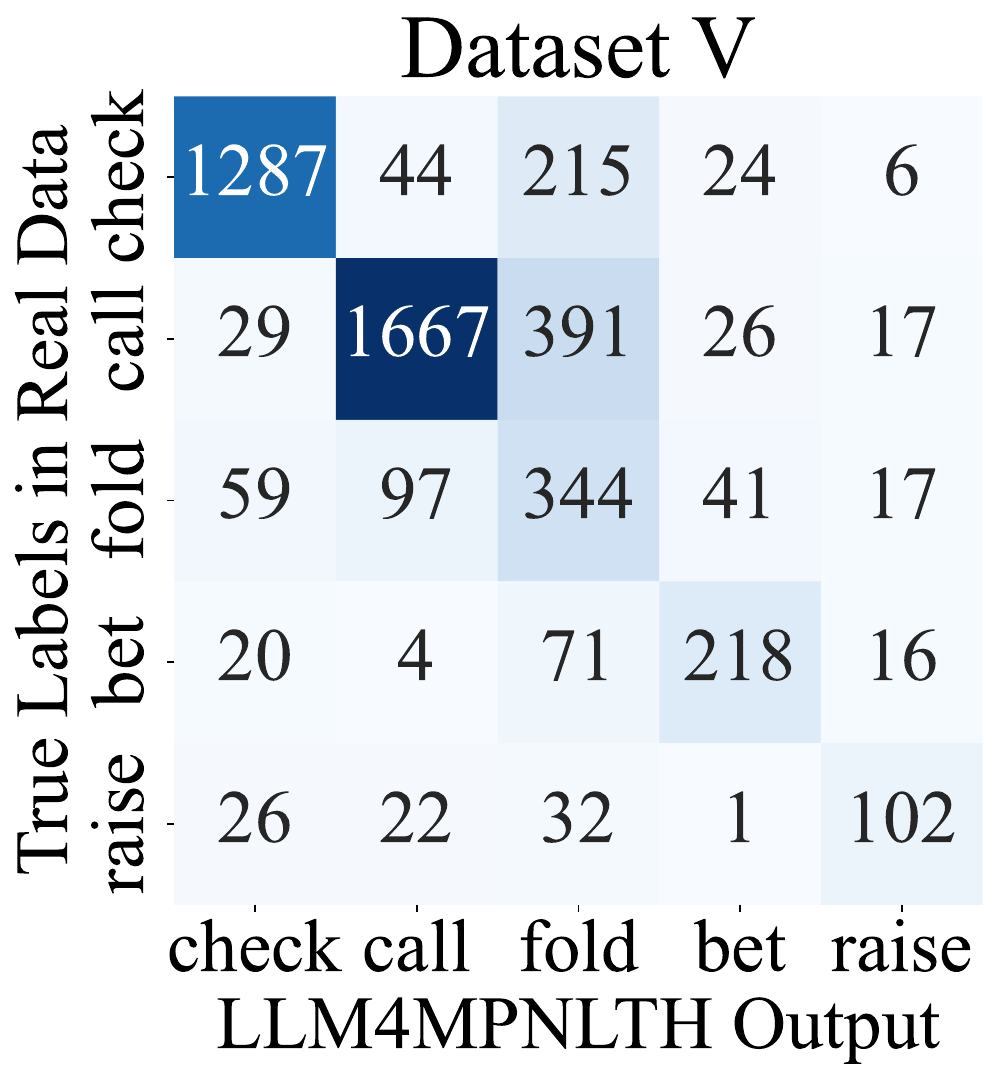}  }
    \hspace{-0.1in}
    \subfigure {
    \includegraphics[width=.24\textwidth]{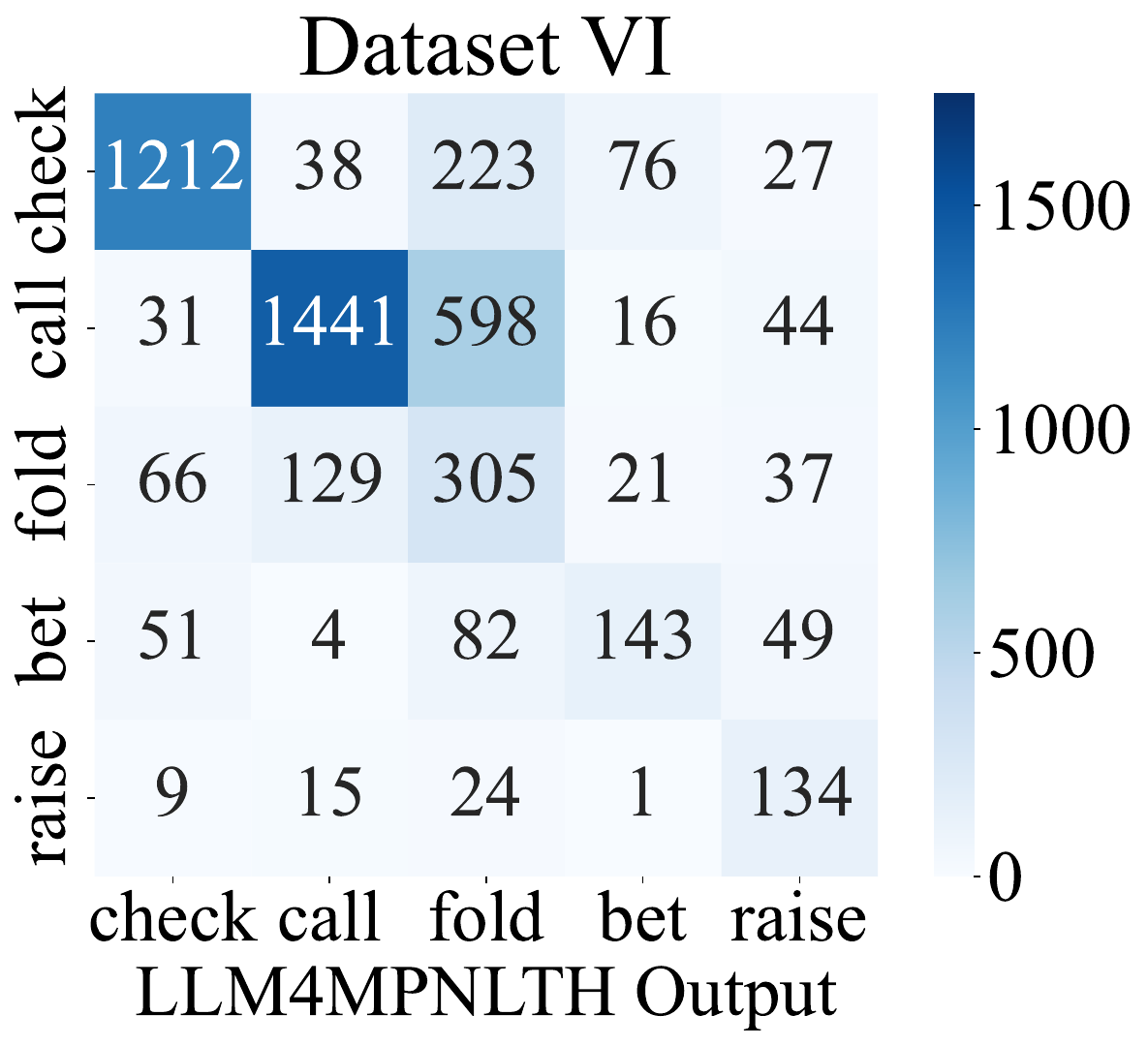}  }
    \captionsetup{width=.99\linewidth}
    \caption{Confusion matrix of test results outputted by PokerGPT fine-tuned using Dataset III to VI. \vspace{-20pt}}
    \label{Fig:heatmap}
\end{figure}


In ablation experiments, we firstly evaluate the effectiveness of each technique used in our model. As shown in Table~\ref{Table_ablation}, by only fine-tuning raw data, Dataset I shows the worst macro F1 score and perplexity, because it is difficult for the LLM to learn effectively from a bunch of unformatted data, whose information is in chaos. Thus, the LLM can not even make actions valid for Poker games. Dataset II shows much better performance than Dataset I, indicating the importance of information filtering applied in our model. After selecting players with different win rate for fine-tuning, we can see there is a trend that the holistic performance of our model is positively related to win rate, where Dataset III has the best performance while Dataset VI is the worst. According to macro F1 score and MSE, fine-tuning on worse dataset leads to not only more incorrect action choices, but also irrational investment. Interestingly, Dataset VI has a similar perplexity to Dataset III to V while performing much worse, indicating it is firm about its bad action choice. This result emphasizes the importance of data quality. In conclusion, all the techniques adopted in this work show effectiveness in different extent, in which prompt selection based on players' win rate presents the strongest effect.

To check the training bias, we show the confusion matrix of classification results of Dataset III to VI in Fig.~\ref{Fig:heatmap}. The test set contains 1000 games with 1576 ``check''s, 2130 ``call''s, 558 ``fold''s, 329 ``bet''s, and 183 ``raise''s. We can see that the model of Dataset III has better prediction on ``check'', ``call'', ``fold'', and ``bet'' than the model of Dataset VI does, indicating that our model can learn strategies more similar to winners' operations from data of high-level players. On the other hand, the reason for the higher accuracy of ``raise'' of Dataset VI can be that the players losing a lot have a stronger ``raise'' inclination which influence the action style of our model. Besides, an interesting appearance is that both the 2 models tend to give more ``fold'' than other actions. This could be attributed to that the LLM has the basic concept of game goal the value judgment of Poker, that is, if it is confusing to choose an action, the model will choose ``fold'' in order to reduce investment. Additionally, compared to Dataset III, the outputs of Dataset VI shows higher ratio of choosing ``raise'', leading to worse macro F1 score, reaffirming the importance of data quality for LLM fine-tuning.

\begin{figure}[htbp]
    \centering
    \includegraphics[width=1\linewidth]{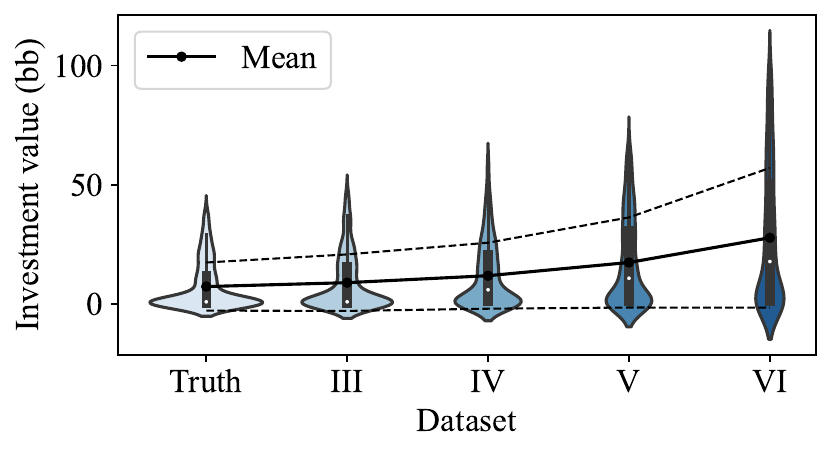}
    \captionsetup{width=.99\linewidth}
    \caption{The distribution of investment values among 1000 games, whose unit is big blind. The 2 dashed lines are the mean with standard deviation. \vspace{-15pt}}
    \label{Fig:investment_distribution}
\end{figure}

Moreover, we record the investment distribution of the truth and predicted investment from models fine-tuned on Dataset III to VI, shown in Fig.~\ref{Fig:investment_distribution}. In this figure, the mean of investment increases and the distribution diversifies with the data quality degrades. Generally, winners intend to invest less than 10 bb for an easy win, while players who win less or lose more incline to invest more money. The results imply that the strategies adopted by high-level players are more conservative, which is similar to winners in the truth. On the other hand, the similarity is, the investment inclination under all cases is to focus on relatively small values, and invest higher values with lower frequency, indicating that our model has learned the common sense of investment risk in games. In terms of the distribution of Dataset VI, the mean of the model's investment even reaches about 10, undertaking much higher risk.

\begin{figure}[htbp]
    \centering
    \includegraphics[width=1\linewidth]{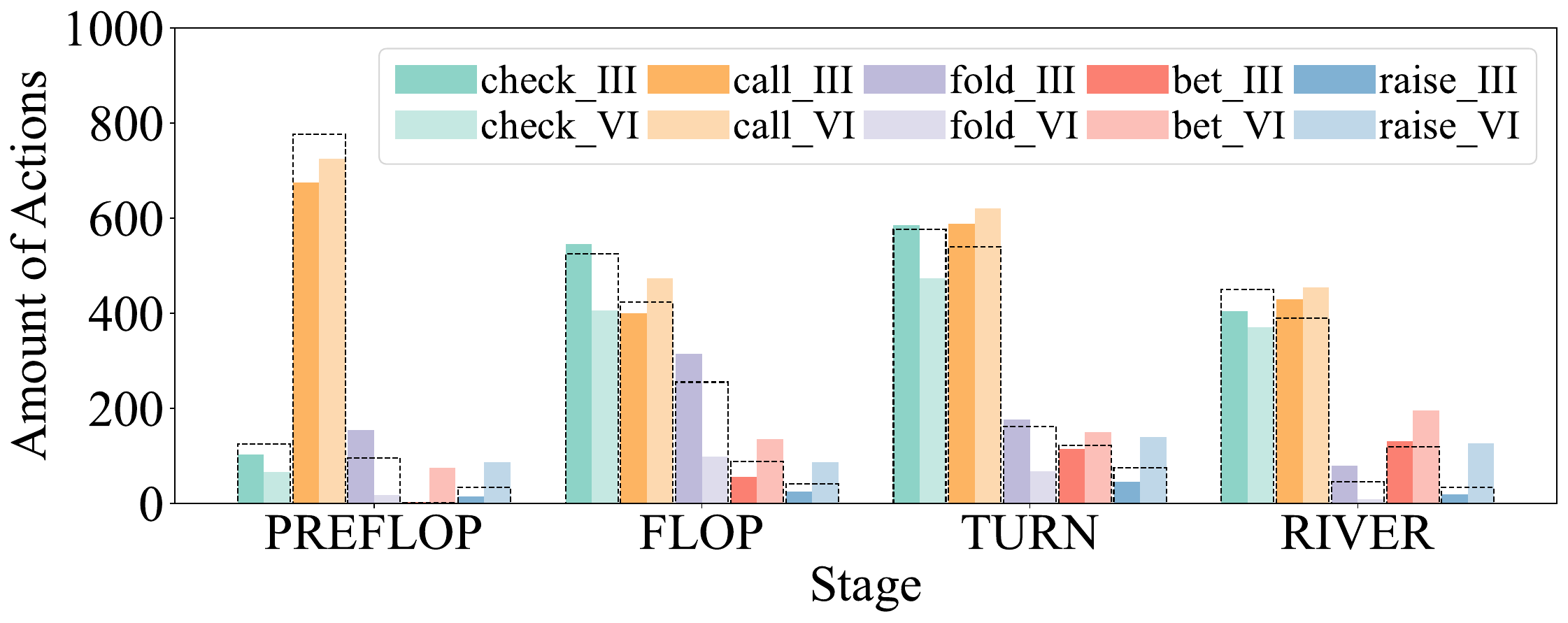}
    \captionsetup{width=.99\linewidth}
    \caption{The predicted actions of 1000 games from LLMs fine-tuned on Dataset III and Dataset VI. The dashed bars represent the real actions. \vspace{-5pt}}
    \label{Fig:phase_label}
\end{figure}

Additionally, we record staged actions of our model fine-tuned on Dataset III and Dataset VI. As shown in~\ref{Fig:phase_label}, the results of PokerGPT on Dataset III always be more likely to ``fold'' in all stages, indicating the model's style is conservative. In contrast, PokerGPT on Dataset VI prone to ``raise'' and ``bet'', and rarely ``fold'', showing its aggressiveness. These 2 phenomenons can explain that players with higher win rates are usually more cautious that those who lose more. Both models tend to ``call'' in later stages, implying that ``call'' is a more general choice in some extent.

As a conclusion, the above ablation experiments validate the effectiveness of our approaches.

\subsection{Multi-Player Evaluation}

\begin{figure}[hbpt]
    \centering   
    \subfigure[]{\label{Fig:player_amount a}
    \includegraphics[width=.2\textwidth]{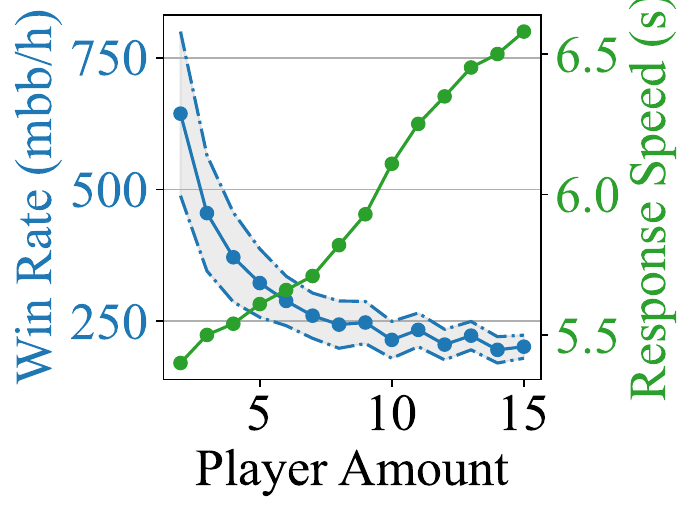}  }
    \hspace{-0.2in}
    \subfigure[]{\label{Fig:player_amount b}
    \includegraphics[width=.26\textwidth]{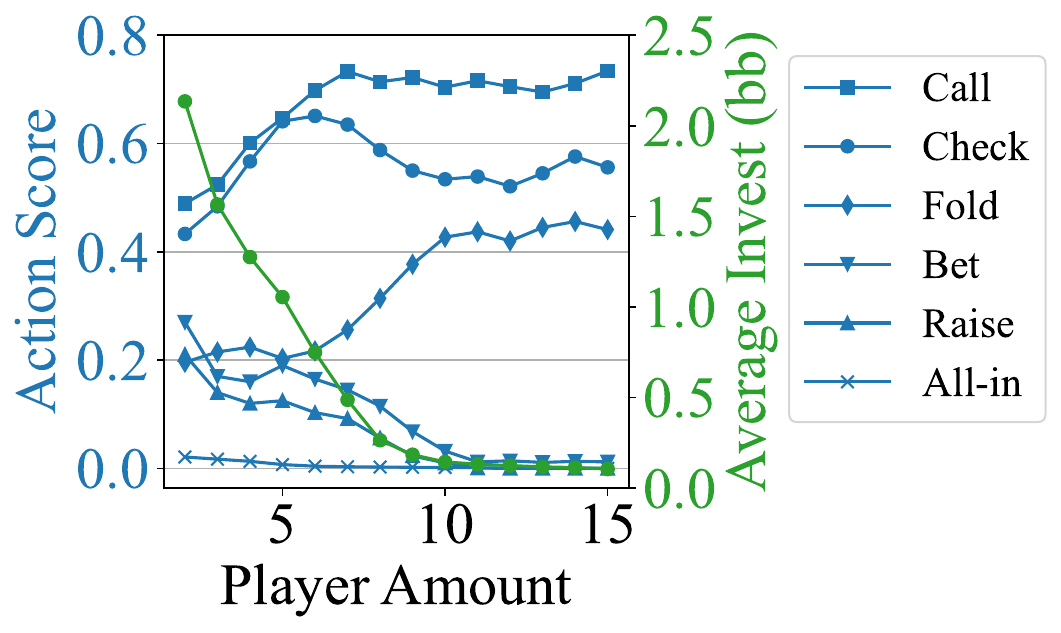}  }
    \captionsetup{width=.99\linewidth}
    \caption{(a) The trends of win rate and response speed when player amount changes. The dashed lines are the standard deviations. (b) Strategy variation during player amount changes. \vspace{-10pt}}
    \label{Fig:player_amount}
\end{figure}

To evaluate the performance of PokerGPT when dealing with multiple players, we make it play 1000 games for each player amount. The Fig.~\ref{Fig:player_amount a} presents the varying tendency of 2 key metrics of our model when players' number increases. The trend of win rate changes more significantly during player amount 2 to 5, then shows a slower declination during player amount from 6 to 9, and finally fluctuations with a small amplitude appears during player amount 10 to 15. When playing 2-player games, our model gets a higher win rate against a rule-based bot which considers equity, indicating the capability of handling the easiest Poker games. Since situations become more complex and the win rate declines when the player amount rises, our model gains less but still positive revenue when competing with more opponents, showing the advantages of our model in multi-player games. 
Meanwhile, the gradual reduction of the shaded part implies that the fluctuation of the win rate decreases. 

Moreover, the response time rises about linearly, while the win rate declines drastically at first and gradually goes moderate. This phenomenon validates the good scalability of our work. Different from CFR, LLMs intend not to fully explore all possible scenarios. Instead, LLMs directly look for connection between inputs and outputs. As the number of players increases, the amount of input information shows roughly linear growth, and the number of possible situations involved in each round of the game consequently goes up. As a result, the linear time increase shows the efficiency of end-to-end learning on LLMs. In conclusion, all these phenomenons demonstrate the advantages of our model in Poker games, and extensive potential in IIGs.

Furthermore, to study the variation of PokerGPT's strategy when facing multiple players, we calculate the action score for each action and the average investment, shown in Fig.~\ref{Fig:player_amount b}. According to action scores, we can observe each action and get insights separately:
\begin{itemize}
    \item \textit{``Call'':} We can see that the score of the ``call'' strategy increases with the player amount, suggesting that the PokerGPT prefers to be more conservative when facing more opponents.
    \item \textit{``Check'':} The score of ``check'' increases till 6-player games, and then deceases into a range. Since ``check'' is not only a compromising behavior but also a bait for aggressive opponents, players can choose ``check'' in any situation, especially in games with small player amount. However, when the scale of games becomes larger than 10 players, bait tactics become less effective. Thus, ``check'' can usually be regarded as a compromise, and chosen in a relatively stable frequency.
    \item \textit{``Fold'':} As ``fold'' score increases a lot when player amount becomes larger than 6, 2 explanations can be drew. When there are more players, it is complex to forecast the rank strength among dozens of hands, leading to a high rate of ``fold''. Also, ``fold'' score can be higher with the game length shortening, since it is reasonable for bots to ``fold'' if possibility of stronger rank existing increases, leading to a quick end of one game.
    \item \textit{``Bet'' and ``Raise'':} The scores of ``bet'' and ``raise'' decrease with the number of players. This implies that the model intends to avoid risk when facing more players. 
    \item \textit{``All-in'':} Actually, ``all-in'' is a special case of ``bet'' or ``raise''. The score of the ``all-in'' is very low and even tends to 0 with more opponents. This reflects that the model rarely chooses this action when player amount increases, as the risk of ``all-in'' is higher.
\end{itemize}

In summary, actions suggested by our model tend to be more cautious and conservative as the number of players increases, which are generally consistent with real games.




\subsection{Analysis of Interaction}


In Fig.~\ref{fig:PokerGPT}, we show the interaction with PokerGPT during a Poker game. In Q1, we ask PokerGPT to give a direct action recommendation and get ``call'' as feedback. Then, in Q2, when we request it to be more aggressive, it suggests to make a ``raise'' to 0.8. It shows that the human language comprehension of LLMs help PokerGPT adjust the strategy according to the user's request. In ``FLOP'' stage with 3 public cards exhibited, PokerGPT successively gives different answers to Q3 and Q4, from ``check'' to ``all-in'', suggesting that PokerGPT has the understanding of card strengths. Besides, in Q6, we ask about Seat 9's rank, with the answer that Seat 9 possibly has a pair or a flush since this player likes to ``raise'' and ``bet'', suggesting that PokerGPT has the awareness of an opponent's strength based on his or her actions. However, from the answer of Q7 and Q9, PokerGPT shows its shortage on guessing cards, probably due to the large quantities of card combinations. Similarly, in Q9, PokerGPT has difficulty of giving a winning probability, since we do not use any probability as a part of the prompt, while LLMs are not proficient in math, neither.


\begin{figure}[htbp]
\centering
\resizebox{\linewidth}{!}{
    \begin{AIbox}{PokerGPT}
    \parbox[t]{\linewidth}{
    {\bf System (constant information):} You are an experienced gambler. Now you need to assist me to make decisions in Texas Hold'em games. You have been provided with a series of observable information:\\
    \small{Player amount: [6], Currency: USD, Blind value: [0.02/0.05], Order: [`2', `3', `5', `6', `7', `9'], Seat 2 is small blind.} \\
    \small{My cards: [`Th', `Ah'], the characteristics of my cards: [``suit'', ``high'', ``close''], My seat: [Seat 2]}}
    \tcbline
    \parbox[t]{\linewidth}{
    \small{\bf Dynamic information:}\\
    \small{{\bf Stage: ``PREFLOP''}, Public cards: [`**' `**' `**' `**' `**']}\\
    \small{My rank: [``High''], Money: [3.92], Action: []}\\
    \small{Seat 3: [`**', `**'], Money: [2.33], Action: [], Discard: [False]}\\
    \small{Seat 5: [`**', `**'], Money: [5.54], Action: [``fold''], Discard: [True]}\\
    \small{Seat 6: [`**', `**'], Money: [3.75], Action: [``fold''], Discard: [True]}\\
    \small{Seat 7: [`**', `**'], Money: [4.22], Action: [``fold''], Discard: [True]}\\
    \small{Seat 9: [`**', `**'], Money: [1.47], Action: [``raises 0.05 to 0.1''], Discard: [False]}\\    
    The pot value is [0.17]\\
    The actions can be: [``fold'', ``raise'', ``call'']. What should I do? If I choose to ``bet'' or ``raise'', then how much? Choose a number from \{(0, 0.05, 0.15, 0.3, 0.5, 1, 2.5, \textit{3.92}\}.
    }
    \tcbline
    {\bf User:} \\
    \parbox[t]{\linewidth}{
    {\bf Q1:}\sethlcolor{green!20}\hl{What should I do?} \\
    {\bf PokerGPT:} \sethlcolor{blue!20}\hl{You should call.}}\\
    
    \parbox[t]{\linewidth}{
    {\bf Q2:} \sethlcolor{green!20}\hl{Please be aggressive.} \\ 
    {\bf PokerGPT:} \sethlcolor{blue!20}\hl{You should raise to 0.5.}}\\
    \tcbline
    {\bf Information updates:}\\
    \parbox[t]{\linewidth}{
    \small{{\sethlcolor{red!20}\hl{\bf Stage: ``FLOP''}}, Public cards: [\sethlcolor{red!20}\hl{`7h' `4h' `2h'} `**' `**']}}\\
    \small{My rank: [\sethlcolor{red!20}\hl{``Flush''}], Money: [\sethlcolor{red!20}\hl{3.82}], Action: [\sethlcolor{red!20}\hl{``call''}]}\\
    \small{Seat 3: [`**', `**'], Money: [\sethlcolor{red!20}\hl{2.23}], Action: [\sethlcolor{red!20}\hl{``call''}], Discard: [False]}\\
    ......\\
    \small{Seat 9: [`**', `**'], Money: [\sethlcolor{red!20}\hl{1.37}], Action: [``raises 0.05 to 0.1''], Discard: [False]}\\    
    The pot value is [\sethlcolor{red!20}\hl{0.3}]\\
    The actions can be: [``fold'', ``check'', ``bet'']. What should I do? If I choose to ``bet'' or ``raise'', then how much? Choose a number from \{(0, 0.05, 0.15, 0.3, 0.5, 1, 2.5, \sethlcolor{red!20}\hl{\textit{3.82}}\}.
    \tcbline
    {\bf User:} \\
    \parbox[t]{\linewidth}{
    {\bf Q3:}\sethlcolor{green!20}\hl{What should I do?} \\
    {\bf PokerGPT:} \sethlcolor{blue!20}\hl{You should check.}}\\
    
    \parbox[t]{\linewidth}{
    {\bf Q4:} \sethlcolor{green!20}\hl{Should I all-in?}\\
    {\bf PokerGPT:} \sethlcolor{blue!20}\hl{As you have a flush, all-in is fine.}}\\
    \tcbline
    {\bf Information updates:}\\
    \parbox[t]{\linewidth}{
    \small{{\bf Stage: ``FLOP''}, Public cards: [`7h' `4h' `2h' `**' `**']}}\\
    \small{My rank: [``Flush''], Money: [3.82], Action: [``call'', \sethlcolor{red!20}\hl{``check''}]}\\
    \small{Seat 3: [`**', `**'], Money: [2.23], Action: [``call'', \sethlcolor{red!20}\hl{``check''}], Discard: [False]}\\
    \small{Seat 9: [`**', `**'], Money: [\sethlcolor{red!20}\hl{1.15}], Action: [``raises 0.05 to 0.1'', \sethlcolor{red!20}\hl{``bets 0.22''}], Discard: [False]}
    The pot value is [\sethlcolor{red!20}\hl{0.52}]\\
    ......
    \end{AIbox}}
    \end{figure}

\begin{figure}[htbp]
\centering
\resizebox{\linewidth}{!}{
    \begin{AIbox}{}
    \small{The actions can be: [``fold'', ``call'', ``raise'']. What should I do? If I choose to ``bet'' or ``raise'', then how much? Choose a number from \{(0, 0.05, 0.15, 0.3, 0.5, 1, 2.5, \textit{3.82}\}.}
    \tcbline
    {\bf User:} \\
    \parbox[t]{\linewidth}{
    {\bf Q5:}\sethlcolor{green!20}\hl{What should I do?} \\
    {\bf PokerGPT:} \sethlcolor{blue!20}\hl{You should raise to 0.5.}}\\
    
    \parbox[t]{\linewidth}{
    {\bf Q6:} \sethlcolor{green!20}\hl{Please guess Seat 9's rank.}\\
    {\bf PokerGPT:} \sethlcolor{blue!20}\hl{Seat 9 likes to raise and bet. Therefore, he might be three of a kind or flush.}}\\

    \parbox[t]{\linewidth}{
    {\bf Q7:} \sethlcolor{green!20}\hl{Please guess Seat 9's cards.}\\
    {\bf PokerGPT:} \sethlcolor{blue!20}\hl{Seat 9 likes to raise and bet. Therefore, he might be there of a kind or flush. But there is no evidence what cards Seat 9 has.}}
    \tcbline
    {\bf Information updates:}\\
    \small{\sethlcolor{red!20}\hl{\bf Stage: ``TURN''}, Public cards: [`7h' `4h' `2h' \sethlcolor{red!20}\hl{`Ks'} `**']}\\
    \small{My rank: [``Flush''], Money: [\sethlcolor{red!20}\hl{3.6}], Action: [``call'', ``check'', \sethlcolor{red!20}\hl{``call''}]}
    \small{Seat 3: [`**', `**'], Money: [2.23], Action: [``call'', ``check'', \sethlcolor{red!20}\hl{``fold''}], Discard: [\sethlcolor{red!20}\hl{True}]}\\
    ......\\ 
    The actions can be: [``fold'', ``check'', ``bet'']. What should I do? If I choose to ``bet'' or ``raise'', then how much? Choose a number from \{(0, 0.05, 0.15, 0.3, 0.5, 1, 2.5, \sethlcolor{red!20}\hl{\textit{3.6}}\}.
    \tcbline
    {\bf User:} \\
    \parbox[t]{\linewidth}{
    {\bf Q8:}\sethlcolor{green!20}\hl{What should I do?} \\
    {\bf PokerGPT:} \sethlcolor{blue!20}\hl{You should bet 1.}}\\

    \parbox[t]{\linewidth}{
    {\bf Q9:} \sethlcolor{green!20}\hl{Please tell me my winning probability.}\\
    {\bf PokerGPT:} \sethlcolor{blue!20}\hl{Sorry, I can't calculate the probability. However, as you have flush, you are very likely to win.}}
    \tcbline
    {\bf Information updates:}\\
    \small{... My Action: [``call'', ``check'', ``call'', \sethlcolor{red!20}\hl{``bet 0.9''}]}\\
    ......\\
    \small{Seat 9: [`**', `**'], Money: [1.37], Action: [``raises 0.05 to 0.1'', ``bets 0.22'', \sethlcolor{red!20}\hl{``fold''}], Discard: [\sethlcolor{red!20}\hl{True}]}\\    
    The pot value is [\sethlcolor{red!20}\hl{1.42}]
    \tcbline
    {\bf \sethlcolor{orange!20}\hl{Player 2 wins \$1.42 with ``Flush''.}}\\
    \end{AIbox}}
    \captionsetup{width=.99\linewidth}
    \caption{The question-answering instance of PokerGPT. \vspace{-10pt}}
    \label{fig:PokerGPT}
    \end{figure}

\section{Conclusion and Future Work}\label{conclusion}
Inspired by the prosperity and prominence of LLMs, our study explores the applicability of LLMs for playing Poker. By conducting prompt engineering on data of real games and fine-tuning a lightweight pre-trained LLM based on RLHF, we have successfully developed an end-to-end solver for Poker with arbitrary amount of players, which is also convenient for interaction. The experimental results are notably satisfactory, showing a great advantage in training cost and inference speed among previous works, while obtaining a competitive win rate. PokerGPT addresses a series of challenges encountered by previous Poker AI, substantiating the considerable potential of LLMs in tackling IIGs.

In subsequent research directions, we aim to further improve the capability of PokerGPT. To empower our model with explainability on decisions, various data should be prepared, including explanations on actions in each round written by humans, probabilities of each action which can be outputted by CFR models, multiple candidate decisions, etc.





\bibliographystyle{IEEEtran}
\bibliography{bib}

\end{document}